\documentclass[sigconf]{acmart}

\usepackage{lipsum,enumitem}
\usepackage{graphicx,subcaption}
\usepackage{array,multirow,arydshln}
\usepackage{algorithm,algpseudocode}

\usepackage{pifont}%
\newcommand{\cmark}{\ding{51}}%
\newcommand{\xmark}{\ding{55}}%

\AtBeginDocument{%
  \providecommand\BibTeX{{%
    \normalfont B\kern-0.5em{\scshape i\kern-0.25em b}\kern-0.8em\TeX}}}

\acmYear{2024}\copyrightyear{2024}
\setcopyright{rightsretained}
\acmConference[MMSys '24]{ACM Multimedia Systems Conference 2024}{April 15--18, 2024}{Bari, Italy}
\acmBooktitle{ACM Multimedia Systems Conference 2024 (MMSys '24), April 15--18, 2024, Bari, Italy}
\acmDOI{10.1145/3625468.3647623}
\acmISBN{979-8-4007-0412-3/24/04}

\newcommand{\ie}[1]{\textit{i.e.}#1}
\newcommand{\eg}[1]{\textit{e.g.}#1}
\newcommand{\mname}[1][ ]{SyMPIE#1}
\begin{document}

\title{A Modular System for Enhanced Robustness of Multimedia Understanding Networks via Deep Parametric Estimation}

\author[Francesco Barbato]{Francesco Barbato\footnotemark}
\email{francesco.barbato@unipd.it}
\orcid{0000-0001-9893-5813}
\affiliation{%
  \institution{University of Padova}
  \department{Samsung Research UK}
  \country{Italy / United Kingdom}
}
\author{Umberto Michieli}
\email{u.michieli@samsung.com}
\orcid{0000-0003-2666-4342}
\affiliation{%
  \institution{Samsung Research UK}
  \country{United Kingdom}
}
\author{Mehmet Kerim Yucel}
\email{m.yucel@samsung.com}
\orcid{0000-0003-1645-0877}
\affiliation{%
  \institution{Samsung Research UK}
  \country{United Kingdom}
}
\author{Pietro Zanuttigh}
\email{zanuttigh@dei.unipd.it}
\orcid{0000-0002-9502-2389}
\affiliation{%
  \institution{University of Padova}
  \country{Italy}
}
\author{Mete Ozay}
\email{m.ozay@samsung.com}
\orcid{0000-0002-7189-7260}
\affiliation{%
  \institution{Samsung Research UK}
  \country{United Kingdom}
}

\begin{abstract}
\setcounter{footnote}{1}
\renewcommand*{\thefootnote}{\fnsymbol{footnote}}
\footnotetext{This work has been produced during an internship at Samsung Research UK.}
\setcounter{footnote}{0}
\renewcommand*{\thefootnote}{\arabic{footnote}}
In multimedia understanding tasks, corrupted samples pose a critical challenge, because when fed to machine learning models they lead to performance degradation. In the past, three groups of approaches have been proposed to handle noisy data: i) enhancer and denoiser modules to improve the quality of the noisy data, ii) data augmentation approaches, and iii) domain adaptation strategies. All the aforementioned approaches come with drawbacks that limit their applicability; the first has high computational costs and requires pairs of clean-corrupted data for training, while the others only allow deployment of the same task/network they were trained on (\ie, when upstream and downstream task/network are the same).
In this paper, we propose \mname to solve these shortcomings. To this end, we design a small, modular, and efficient (just 2GFLOPs to process a Full HD image) system to enhance input data for robust downstream multimedia understanding with minimal computational cost. Our \mname is pre-trained on an upstream task/network that should not match the downstream ones and does not need paired clean-corrupted samples.
Our key insight is that most input corruptions found in real-world tasks can be modeled through global operations on color channels of images or spatial filters with small kernels.
We validate our approach on multiple datasets and tasks, such as image classification (on ImageNetC, ImageNetC-Bar, VizWiz, and a newly proposed mixed corruption benchmark named ImageNetC-mixed) and semantic segmentation (on Cityscapes, ACDC, and DarkZurich) with consistent improvements of about 5\% relative accuracy gain across the board.\footnote{Our codebase and the new ImageNetC-mixed benchmark are available at\\\url{https://github.com/SamsungLabs/SyMPIE}.}
The code of our approach and the new ImageNetC-mixed benchmark will be made available upon publication. 

\begin{figure}[t]
    \centering
    \includegraphics[width=\linewidth]{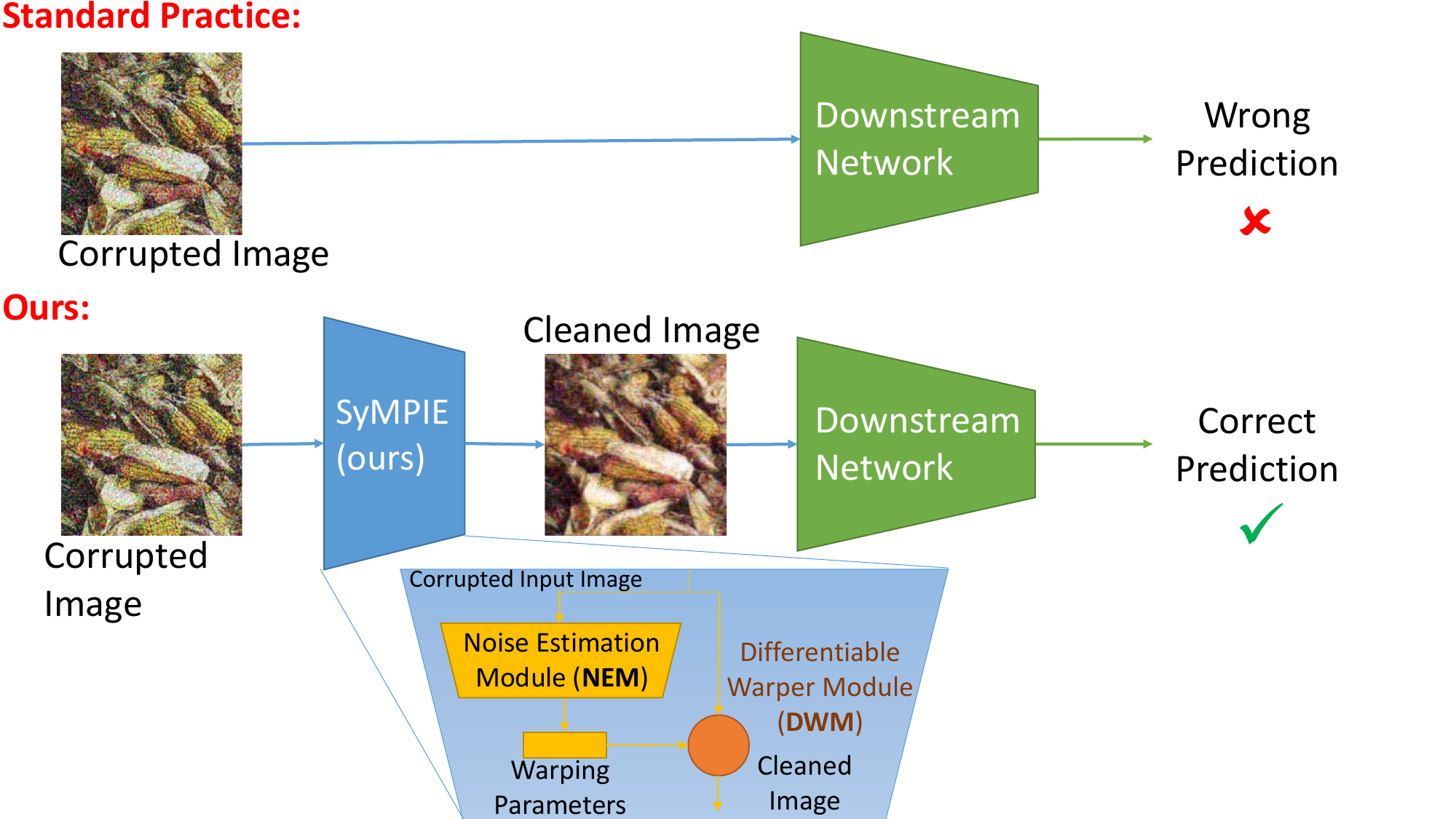}
    \caption{An illustration of our modular system (\mname[]) for efficient image enhancement targeting increased model robustness to corruptions in different multimedia tasks. \mname contains two modules, namely, NEM and DWM (see Fig.~\ref{fig:warper} for their details). \mname estimates parameters to clean input samples and can be integrated into any deep network architecture for multimedia understanding. \mname is fully differentiable and can be trained end-to-end on an upstream task, without the need for paired clean/corrupted images to enhance images automatically. We use \textit{upstream task} to identify the one used to pre-train our system (\eg, classification) and \textit{downstream task} to identify the deployment one.} 
    \label{fig:abstract}
\end{figure}

\end{abstract}



\maketitle

\section{Introduction} 
\label{sec:introduction}
Deep learning models have been  widely used in several multimedia systems and applications.
Nonetheless, recent studies have explored their robustness to corrupted input data~\cite{hendrycks2019robustness,Hendrycks_2021_ICCV}, showing critical performance degradation affecting the quality of experience of many downstream applications such as extended reality~\cite{albreiki2022robustness} and video streaming~\cite{wang2023test}.
Indeed, the data-hungry nature of deep neural network (DNN) models, as well as the ever-growing complexity of network architectures, make the generated models sensitive to even mild distribution shifts causing a severe degradation in performance~\cite{recht2019imagenet,testolina2023selma,rizzoli2023syndrone}.
These eventualities are often naturally met in many real-world applications, where data may unavoidably encounter natural alterations or corruptions~\cite{hendrycks2019robustness,michieli2023online}, such as sensor degradation (\textit{e.g.}, \textit{shot noise}, \textit{defocus blur}) or compression artifacts, to mention a few.

Considering the increasing adoption of deep models, this issue has become of paramount relevance.
Therefore, a new research field emerged to attempt to make models more robust under different perspectives.
Existing approaches to improve model robustness can be categorized into three main branches: (i) data augmentation approaches, (ii) test-time adaptation approaches, and (iii) enhancing and denoising approaches, such as autoencoder-based, generative, and adversarial denoisers (\eg, GANs, diffusion models, \textit{etc.}). 

Data augmentation techniques~\cite{yucel2023hybridaugment++,Chen_2021_ICCV,hendrycks2020augmix} either re-train AI models from scratch or fine-tune them by applying  a large set of general-purpose augmentations (often synthetic) that mimic common corruptions encountered in the real world on data. 
In other words, this paradigm aims to produce a model that is robust against corrupted images and that can be very effective in improving model abilities to generalize under data distribution shift.
The main drawbacks of this approach are that the trained model can only perform the task that has been trained with supervision on; the trained model tends to perform well on images with training-like distortions;
the types of augmentations should be decided \textit{a priori}, and they cannot be easily tuned to a scenario with variable corruptions (unless via expensive retraining).

To improve model robustness to variable-type corruptions at inference time compared to training time, Test-Time Adaptation (TTA) methods have been proposed~\cite{Hu2021MixNormTA,Khurana2021SITASI,Zhang2021MEMOTT}.
TTA methods focus on resolving data distribution shifts directly at test-time via dynamic updates of pre-trained models based on the specific characteristics of the target test data~\cite{gong2022note,wang2021tent,niu2023towards}. 
TTA methods enable a model to improve its performance when encountering variations, unseen examples, or changing conditions at test time. Unlike data augmentation techniques, TTA methods do not involve expensive training procedures, but 
they are still bounded by the same task to be used during the pre-training and testing/adaptation phases.

Other popular techniques are enhancement and denoising, \ie, the process of improving input samples before they are fed to the downstream network~\cite{tian2020deep}.
The most promising techniques are nowadays based on deep learning models. For instance, autoencoders are used to encode compact representations of  input samples and then decode them to minimize a reconstruction loss with respect to ground truth samples~\cite{akbari2020generalized,zhang2023color,LORE2017650}. Generative and/or adversarial models (\eg, diffusion models~\cite{ulhaq2022efficient,dhariwal2021diffusion,Yi_2023_ICCV} or GANs~\cite{chen2020reusing,wang2020gan,li2022spn2d,nie2021gigan,9204448}) modify input samples by iterative processes or style transfer techniques.
The main drawbacks of these methods are the need for expensive inference time and for paired clean-corrupted samples at training time while being able to handle in-domain corruptions only (\ie, particular corruption types seen at training time).
        
Overall, current  techniques all come with some key limitations that we briefly discussed.
In this paper, we propose a System for Modular Parametric Image Enhancement (\mname[]), to overcome some of these  limitations.
\mname implements a hybrid strategy combining the best of data augmentation and enhancement approaches while being computationally efficient and fully compatible with any model and any downstream task. 

We propose to tackle the issue from a completely different direction, designing a modular system that predicts parameters used by an ad-hoc module to enhance the received input (see Fig.~\ref{fig:abstract}). 
We utilize generic data augmentation pipelines and common upstream task/network (e.g., image classification with ResNet50) to pre-train our model (System for Modular Parametric Image
Enhancement, SyMPIE).
Our key insight is that a large portion of input corruptions found in the real world can be modeled by  applying either global operations on color channels (for example, a night scene can be approximated by a darkened daytime scene), or  spatial noise which can be filtered out by a fast convolution operation with small kernel size (for example, a Laplacian filter or a blurring operation can be used to sharpen an image or to reduce noise on the image).
Additionally, our model is pre-trained minimizing a standard classification task loss, alleviating the need for paired clean-corrupted samples during pre-training. In other words, the downstream task and model do not necessarily need to match the upstream task and model used for pre-training our model. 
For example, we may train our model on a corrupted ImageNet~\cite{russakovsky2015imagenet,hendrycks2019benchmarking} benchmark for image classification using a Convolutional Neural Network (CNN), and deploy it on semantic segmentation benchmarks~\cite{SDV21} with adverse conditions to support a transformer architecture.

Therefore, our \mname is similar to enhancement and denoising approaches, with the key advantage of containing much smaller modules that are pre-trained without paired clean and corrupted ground truth.

Our contributions can be summarized as follows.
\begin{itemize}
\item We propose a lightweight modular image enhancement system, named \mname[], which predicts parameters of \textit{ad-hoc} operators that are applied to input samples to improve the content understanding. 

\item In the experimental analyses, \mname consistently improves the accuracy of  downstream models for the corresponding task with minimal impact on the number of compute operations (about 2GFLOPs on top of 343GFLOPs of a classical ResNet50 architecture for a full HD resolution input) and can be seamlessly applied on top of any competing approach improving their accuracy (up to 8\% relative gain) without re-training.
Unlike data augmentation methods, \mname does not require long training procedures. 

\item Unlike existing denoising approaches, \mname is fast, does not require paired clean-corrupted samples during pre-training, and can be reused in multiple setups without losing its efficacy. In particular, \mname proves to be fully compatible with any convolutional and transformer-based architecture, including recent foundation models (\eg, CLIP). \mname enhances the accuracy of the aforementioned models on several downstream tasks (\eg, image classification, semantic segmentation) in the presence of corrupted test data improving the quality of experience of the final end users.
\end{itemize}

\section{Related Works}
\textbf{Robust generalization.} The robustness of computer vision models against distribution shifts is of vital importance in multimedia applications. Starting with the discovery of adversarial examples~\cite{szegedy2013intriguing}, many different robustness venues have been explored by the community, such as common image corruptions~\cite{hendrycks2019benchmarking}, images with conflicting  shapes/textures~\cite{geirhos2018imagenet}, and style variations~\cite{lin2021can}. The core idea behind robust generalization is to make multimedia models increasingly invariant to such shifts; \textit{e.g.} a model trained on a training set should \textit{generalize} well to samples with unseen styles, corruptions and perturbations.
Arguably, the most common effects found on real data are image corruptions, which have been standardized and categorized in special benchmark datasets~\cite{hendrycks2019benchmarking} for the evaluation of deep learning architectures. This is the scenario we focus on in our paper, due to the practical importance.

As mentioned in Section \ref{sec:introduction}, there are various methods that aim to provide robustness to computer vision models. These methods can be categorized into three groups; i) data augmentation methods, ii) test-time adaptation methods, and iii) preprocessing (\eg,  denoising and enhancement) methods. 

\noindent \textbf{Data augmentation} methods are built on a simple premise; one can achieve robustness by retraining or fine-tuning a model on a training set updated with the samples that the original model failed on (\textit{e.g.}, unseen distributions). Adversarial training~\cite{madry2017towards} has been one of the most prominent examples of such methods, where adversarially corrupted examples are included in the training set during retraining of the model. Since then, the same core idea has been used by a plethora of methods that addressed robustness from a data augmentation perspective. AugMix and its variants~\cite{wang2021augmax,hendrycks2020augmix} proposed to use randomly sampled augmentations in a cascaded manner to diversify the training distribution. DeepAugment proposed to use image translation models to produce new training images~\cite{hendrycks2021many}. Learned augmentation policies tailored for improved robust generalization are proposed in~\cite{cubuk2018autoaugment}. The use of fractals and feature maps to create new samples~\cite{hendrycks2022pixmix}, frequency-specific expert model ensembles~\cite{saikia2021improving}, max-entropy augmentations~\cite{modas2022prime} and frequency-spectrum perturbations~\cite{Chen_2021_ICCV,yucel2023hybridaugment++} are some examples of data augmentation methods which have proven themselves successful against combating the issues caused by distribution shifts. 

\noindent \textbf{Test-time adaptation.} Although they currently hold the state-of-the-art on many benchmarks, data augmentation methods have a key disadvantage, where they require retraining or finetuning every time they encounter (and fail against) new distributions, which may or may not be unknown in practical scenarios. Assuming one can perform these expensive updates periodically, model capacity issues as well as catastrophic forgetting~\cite{kirkpatrick2017overcoming} are likely to be new accuracy bottlenecks, which will inevitably lead to practical issues in deployment scenarios.

In order to avoid such problems, another branch of methods focuses on analyzing input images to detect whether they are of \textit{unseen} distribution. Using network prediction consistency~\cite{lu2017safetynet,feinman2017detecting}, specialized detector models~\cite{grosse2017statistical} and sample statistics~\cite{xu2017feature} have proven to be successful for adversarial examples, and similar ideas have been extended to common image corruptions~\cite{Hu2021MixNormTA,Khurana2021SITASI,Zhang2021MEMOTT,gong2022note,wang2021tent,niu2023towards}. 

Going further than just detecting critical samples, more recent methods aim to improve robustness in test-time. Tent~\cite{wang2021tent} minimizes the test-time model prediction entropy via optimizing channel-wise affine parameters per-batch. MixNorm~\cite{Hu2021MixNormTA} points out the inherent assumption of having samples from a single distribution in a batch, and its detriment on models with batch-norm layers. They then propose MixNorm layers, a replacement for batch-norm layers during test-time that adapts to query image statistics by using novel views of the same image. MEMO~\cite{Zhang2021MEMOTT} minimizes the entropy of predictions of the input sample point and its augmented views, which improves robustness even with a single image. NOTE~\cite{gong2022note} proposes instance-aware batch-norm layers to improve robustness with a single image without the overhead of additional forward passes (\textit{e.g.}, for augmented views of the input sample) required for other methods. SAR~\cite{niu2023towards} proposes a new entropy minimization method that takes into account samples with large gradients and avoids collapsing to trivial solutions, thus improving overall robustness performance. 

\noindent \textbf{Preprocessing.} Test-time adaptation methods largely alleviate the issues inherent to data augmentation approaches; they perform partial updates at most, therefore they are fast and cheap. However, they still require a form of training, which requires periodic updates in deployment scenarios where data shift is ever-present.

Another branch of methods that aims to achieve robust generalization can be roughly categorized as preprocessing methods. These approaches do not tackle the problem as a normalization issue, unlike test-time adaptation methods. Instead, they attempt to bridge the distribution gap between the query sample and the model training set, by trying to \textit{enhance} and \textit{denoise} the query image. Some naive preprocessing strategies such as label smoothing, spatial smoothing, and total-variance minimization are shown to help improve robustness~\cite{yucel2022robust,xu2017feature}. Thinking of the corruption over the image as noise, conventional auto-encoding mechanisms can be used to purge the input image of its corruption-induced elements~\cite{akbari2020generalized}. Essentially, these methods aim to learn an approximation of a transformation over the query image, which makes the underlying issue an image-to-image translation problem. There are many image translation methods using GANs~\cite{chen2020reusing,wang2020gan,li2022spn2d,nie2021gigan} or more powerful diffusion models~\cite{ulhaq2022efficient,dhariwal2021diffusion} that can be either used off-the-shelf or updated to work well with our scenario. Although these methods do not necessarily update the actual downstream task model, they still require per-distribution updates for the model, which makes it expensive. Furthermore, the enhancement or denoising model adds an expensive step that might not be feasible for certain deployment scenarios.

The proposed approach aims to melt the best characteristics of these methods in a single pot: high accuracy, fast execution time, cross-distribution and cross-task generalization, and removing the need for paired data for training.

\section{A Modular System for Improving Model Robustness}
In this section, we introduce the main contributions of this work. 
We detail our two-module network architecture in Sec.~\ref{sub:ours}. Its training procedure exploits exponentially smoothed versions of our modules for improved generalization and is discussed in Sec.~\ref{sub:training}. 
Finally, the inference process is described in Sec.~\ref{sub:inference}.

\subsection{A System for Modular Parametric Image Enhancement (\mname[])}\label{sub:ours}
Our System for Modular Parametric Image Enhancement (\mname[]) is made of two modules.
The first module (Sec.~\ref{subsub:colm}) is the Noise Estimation Module (NEM), $E$, which is implemented by a small CNN to  estimate a set of parameters used by the second module from training samples.
The second module (Sec.~\ref{subsub:colw}) is the Differentiable Warping Module (DWM), $D$, and is used to process input samples in order to remove distortions from them.
To obtain the final \mname architecture, we combine the two modules in a single block by $U = D \circ E$ where $\circ$ denotes the module composition.

\begin{figure}
    \centering
    \includegraphics[trim=0cm 7.5cm 15.7cm 0cm,clip,width=\linewidth]{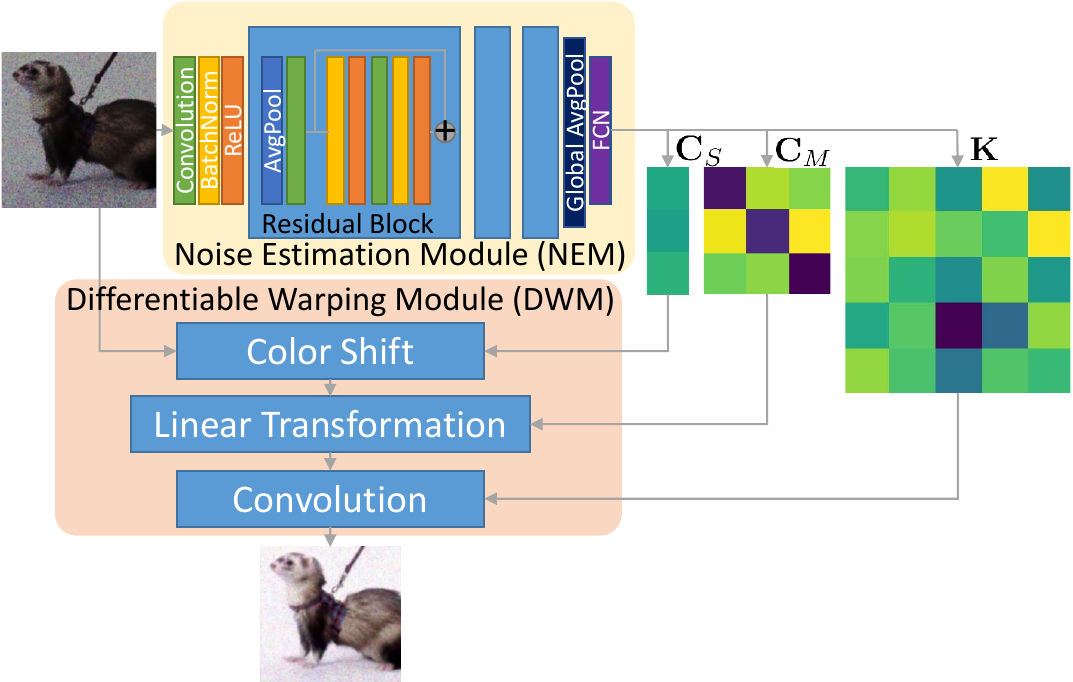}
    \caption{A detailed scheme of our modules working together to enhance the content of an image. The Noise Estimation Module (NEM) receives a corrupted input and predicts a triple of parameters $(\mathbf{C}_S, \mathbf{C}_M, \mathbf{K})$. These parameters are used by the Differentiable Warping Module (DWM) to enhance the image using parametric operators.}
    \label{fig:warper}
\end{figure}

A detailed overview of our modules is shown in Fig.~\ref{fig:warper}:
 \mname can be inserted between the input data and any generic downstream model.
Three key features distinguish our approach from existing image enhancement approaches, namely: (i) the removal of paired clean-corrupted images as a requirement for training; (ii) the prediction of enhancing %
parameters rather than of the cleaned images directly; (iii) the ability to generalize across several upstream and downstream tasks and networks.

Existing works, indeed, train enhancer models as a regression task on the clean images, effectively optimizing the reconstruction PSNR (Peak Signal-to-Noise Ratio). We tackle the issue from a completely different point of view, focusing on efficiency and downstream accuracy, instead. 
Incidentally, this allows our model to be trained and fine-tuned on any available dataset, especially on real-world datasets that contain naturally corrupted images with no clean counterpart ground truth. 

\subsubsection{Noise Estimation Module (NEM)}\label{subsub:colm}

To estimate the parameters needed for enhancement, we designed a small residual CNN, $E$, which estimates explicitly the set of parameters used by the differential warping module $D$.
Motivated by the observation that most of the distortions of the input samples affect the color space only~\cite{mintun2021interaction,MASOOD2014107,zheng1993noise}, we identify a set of filters and linear transformations to be applied to the input samples for their improvement.

In our setup, we aim at estimating the parameters of i) a filter kernel $\textbf{K} \in \mathbb{R}^{K \times K}$ (where kernel size $K$ used in the experiments is given in Sec.~\ref{sub:impl}); ii) a linear transformation of color channels, represented by the matrix $\mathbf{C}_{M} \in \mathbb{R}^{d\times d}$; and iii) a global color shift $\mathbf{C}_{S} \in \mathbb{R}^{d}$ where $d$ is the number of input channels (\eg, $d=3$ for RGB images).
This set of parameters allows our architecture to model several naturally occurring corruptions, such as under/over-exposed images, sensor noise (\eg, Gaussian, impulse, shot noise), unbalanced white point, \textit{etc.} without the need to employ expensive deep learning models for enhancement.

To estimate the above parameters and keep the additional footprint minimal, we implement the NEM as a small residual CNN (see the top half of Fig.~\ref{fig:warper}). 
This allows the NEM module to efficiently extract global information from an input image, and predict suitable parameters for the subsequent module.
More in detail, $E: \mathcal{X} \ni \mathbf{X}  \mapsto (\textbf{K}, \mathbf{C}_{M}, \mathbf{C}_{S}) \in \mathbb{R}^{A}$, where $\mathbf{X}$ is a corrupted input, $\mathcal{X}$ denotes the space of input images, and $(\textbf{K}, \mathbf{C}_{M}, \mathbf{C}_{S})$ are the parameters used for the cleaning process.
More implementation details are given in Sec.~\ref{sub:impl}.

Moreover, in the vanilla deep learning-based image enhancement formulation \cite{akbari2020generalized,LORE2017650}, the network weights used to process the images are fixed. Therefore, the models learn to partition their own parameter spaces to deal with different situations, reducing the overall efficacy and parameter efficiency.
Our model, instead, explicitly predicts the parameters depending on the input, therefore, our model can seamlessly adapt to multiple types of corruptions without the need to partition its parameter space. 

\subsubsection{Differentiable Warping Module (DWM)}
\label{subsub:colw}
To enhance input images using predictions obtained from the NEM, we use our Differentiable Warping Module (DWM), $D$, depicted  in the bottom half of Fig.~\ref{fig:warper}. The module receives as input a corrupted image and the parameters estimated by the NEM and applies the parametric operators to the former to enhance the image. Formally, we have ${D: (\mathbf{X}, (\textbf{K}, \mathbf{C}_{M}, \mathbf{C}_{S})) \mapsto \bar{\mathbf{X}}}$, where $\bar{\mathbf{X}} \in \mathcal{X}$ is the improved image.

The operations of $D$ are completely differentiable, allowing backpropagation of gradients through the DWM, and optimizing the predictions of the NEM in an unsupervised manner. This enables our method to be trained without any paired data, since gradients propagated through the DWM come directly from predictions over a frozen upstream model. In other words, the NEM is trained to enhance samples in order to maximize the performance of a frozen upstream network.
This is a great advantage compared to existing works which either cannot enhance the inputs to improve the performance of downstream models (\eg, diffusion models) or cannot effectively leverage the gradient flow to produce cleaner inputs (\eg, adversarial approaches).
Furthermore, when modifying input samples directly, several approaches~\cite{miller2020adversarial} introduce distortions that appear random and unnoticeable to the human eye but can completely change the output prediction of the model.
In our setup, we avoid this by directly estimating the parameters of transformations whose kernels are translation-invariant (that is, given a translation $T$, the parameters are generated by a multidimensional function $\vec{\pi}(x) = \vec{\pi}(T(x))$ for all pixel locations $x$).
In particular, our modules apply transformations to the whole image at the same time rather than multiple transformations on local subsets of pixels. Therefore, our modules enforce a stronger and visible change in the input image to allow a change in the downstream prediction.
In general, our approach is optimized to enhance the content of an input and improve the performance of the upstream task,
not just the perceived image quality, unlike the other approaches.
The use of global operations also stabilizes the gradients received by the NEM, enabling its training with fewer samples than other approaches (\eg, when compared to augmentation strategies, our approach requires up to 60 times fewer training samples).

\subsection{Training Process}
\label{sub:training}

\begin{figure}[t]
    \centering
    \includegraphics[trim=0cm 5.7cm 3.9cm 0cm,clip,width=\linewidth]{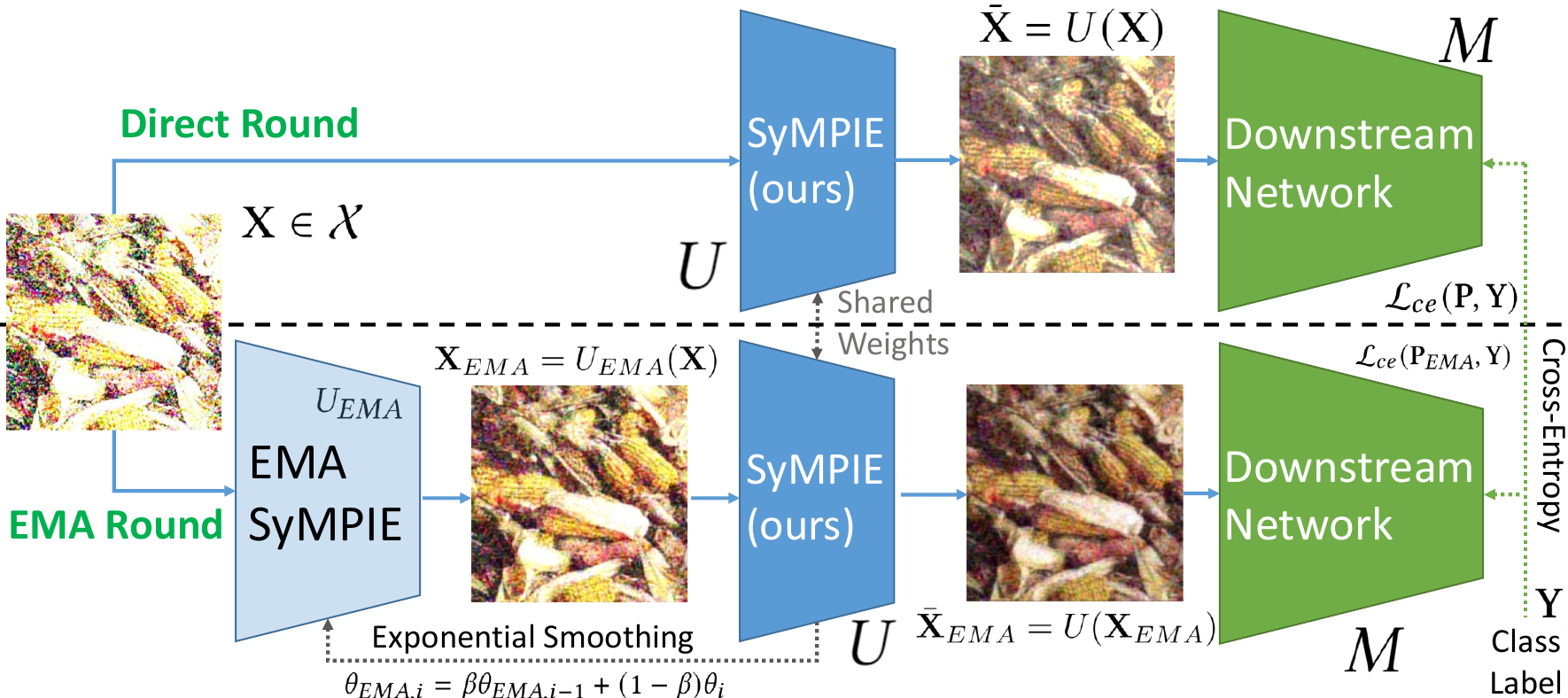}
    \caption{An overview of the training procedure of our modular system.}
    \label{fig:training}
\end{figure}

\begin{algorithm}[t]
    \caption{Training step of \mname[]. 
    }
    \label{alg:training}
    \begin{algorithmic}
    \Require $\mathbf{X} \in \mathcal{X}$ is an input image, and $\mathbf{Y} \in \mathcal{Y}$ is a vector of the corresponding ground truth category label.
    \Require $M$ is the pre-trained upstream task model, $U$ is our \mname to be optimized and $U_{EMA}$ is its smoothed version.
    \Require $\beta$ is the exponential smoothing rate, and $\lambda_{EMA}$ is the weight of the regularization loss.
    \State $ \mathbf{X}_{EMA} \gets U_{EMA}(\mathbf{X})$.
    \State $ \bar{\mathbf{X}} \gets U(\mathbf{X})$.
    \State $ \bar{\mathbf{X}}_{EMA} \gets U(\mathbf{X}_{EMA})$.
    \State $ \mathbf{P} = M(\bar{\mathbf{X}}) $.
    \State $ \mathbf{P}_{EMA} = M(\bar{\mathbf{X}}_{EMA}) $.
    \State $ l_1 = \mathcal{L}_{ce}(\mathbf{P}, \mathbf{Y})$.
    \State $ l_2 = \mathcal{L}_{ce}(\mathbf{P}_{EMA}, \mathbf{Y})$.
    \State Backpropagate the aggregated loss $l = l_1 + \lambda_{EMA} l_2$.
    \State Update the parameters $\theta$ of $U$.
    \State Update the parameters of $U_{EMA}$ by $\theta_{EMA} \gets \beta \theta_{EMA} + (1-\beta) \theta$.
    \end{algorithmic}
\end{algorithm}

A key benefit of our module is its end-to-end training on any given upstream task (we train it on image classification tasks in our experiments) for its deployment on any downstream task (\eg, image classification, semantic segmentation, \textit{etc.}) without further need of fine-tuning it. 
In the following, we describe the training procedure used to optimize our model (summarized in Fig.~\ref{fig:training}) and the data augmentation strategy used during the tuning. A detailed description is provided as pseudo-code in Algorithm~\ref{alg:training}.

We use a frozen model $M$ pre-trained on a source upstream task (\eg, classification) to provide class predictions from the cleaned input samples ($\mathbf{P} = M(\bar{\mathbf{X}})$), and we compare them to the labels $\mathbf{Y} \in \mathcal{Y}$ using the cross-entropy loss $\mathcal{L}_{ce}$, to obtain $l_1=\mathcal{L}_{ce}(\mathbf{P}, \mathbf{Y})$.

As will be shown in Sec.~\ref{subsub:iter}, a common failing point of denoisers is modal collapse upon their iterative application to the same image.
To avoid this, we introduce an additional regularization term to the training loop, which employs the exponentially smoothed version of our modules (we call this $U_{EMA}$). 
The parameters of $U_{EMA}$ at each training iteration $i>0$ are computed by $\theta_{EMA, i} = \beta \theta_{EMA, i-1} + (1-\beta) \theta_{i}$, where $\beta$ is the exponential smoothing rate and $\theta_{EMA, 0}=\theta_{0}$ with $\theta_{i}$ being the parameters of $U$ at iteration $i$. 
During the optimization, $U_{EMA}$ is used to generate an intermediate image $\mathbf{X}_{EMA} = U_{EMA}(\mathbf{X})$, which will show different visual cues than $\mathbf{X}$, aiding in the generalization. 

The intermediate sample is then fed to the modules currently being optimized (\ie, $U$) to obtain $\bar{\textbf{X}}_{EMA} = U(\mathbf{X}_{EMA})$, which is used in the computation of the regularization loss term. 
The two-step sample is then processed as normal, obtaining a prediction from the classifier model $\mathbf{P}_{EMA} = M(\bar{\mathbf{X}}_{EMA})$ to compute a new loss term $l_2=\mathcal{L}_{ce}(\mathbf{P}_{EMA},\mathbf{Y})$, weighted by $\lambda_{EMA}$.

Our modules attempt to enhance input samples which could not otherwise be handled effectively by popular downstream networks. 
For this purpose, we design a set of augmentations that serves as a proxy for the distortions experienced at deployment time, as we describe next.

\subsubsection{Data Augmentation Pipeline}
\label{subsub:color}

To train our modules, we employ a variegate data augmentation pipeline, serving two main purposes: 
(i) to encourage generalization of our architecture over a vast array of possible input corruptions, 
(ii) to reflect conditions that are seen in the real world during deployment.
In particular, we trained our architecture on the ImageNet-1k~\cite{russakovsky2015imagenet} dataset augmented using the corruptions proposed in~\cite{hendrycks2019benchmarking} together with four additional ones to mimic adverse weather conditions at variable degree of severity. 

The four additional corruptions are the following:
\setlist{nolistsep}
\begin{enumerate}[nosep]

\item The \textit{darken} corruption mimics the effect of under-exposure of the scene by reducing the intensities of all pixels in a consistent fashion. For example, this situation may happen after encountering glare that forces cameras to reduce the exposure before re-adjusting. 
A visual example of the resulting images can be seen in the first row of Fig.~\ref{fig:corrupions}.

\begin{figure}[t]
    \centering
    \includegraphics[trim=0cm 0cm 0cm 0cm,clip,width=\linewidth]{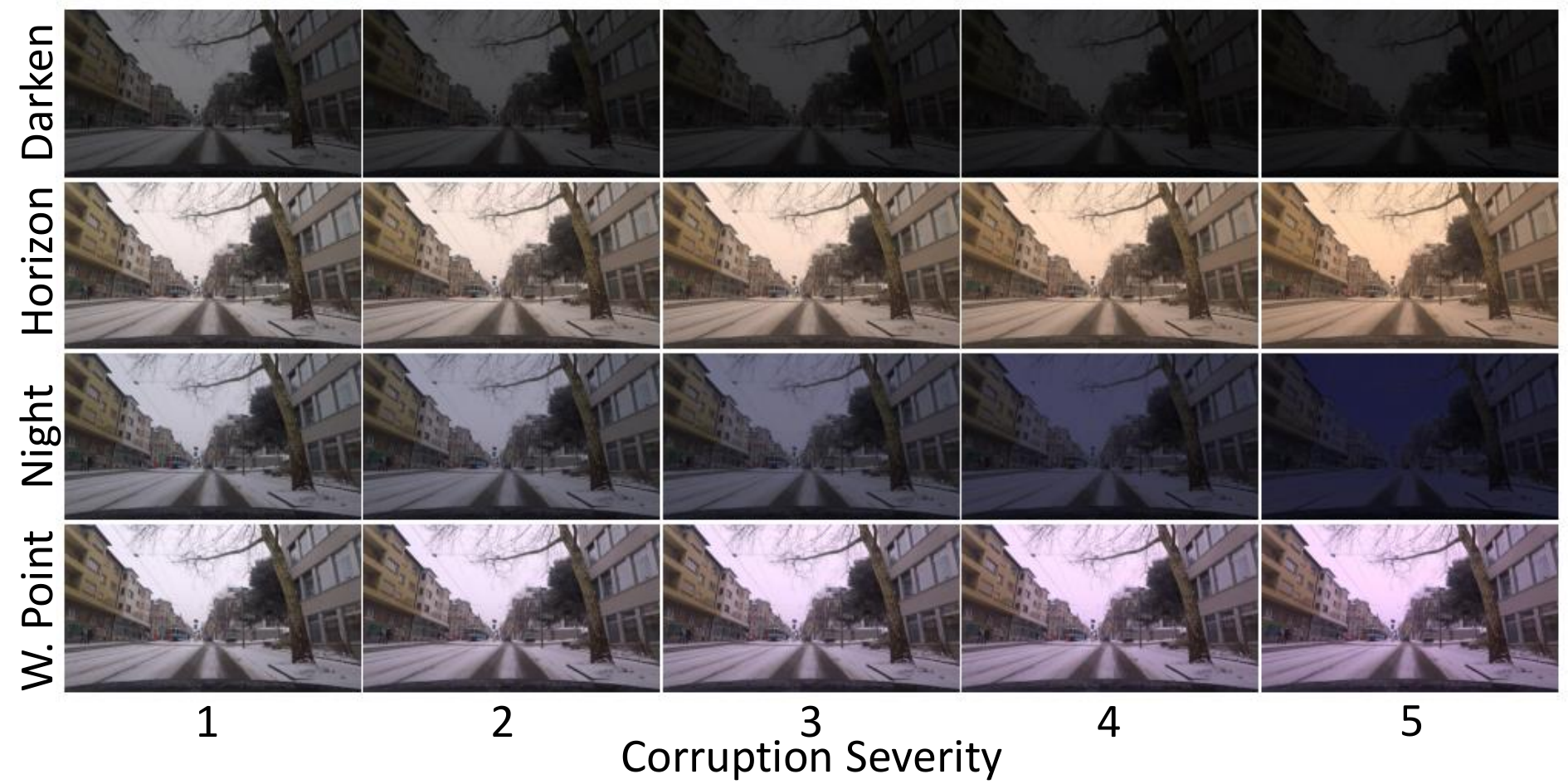}
    \caption{Qualitative examples of our new input corruptions.}
    \label{fig:corrupions}
\end{figure}

\begin{figure}[t]
    \centering
    \begin{subfigure}{.45\textwidth}
        \begin{subfigure}{.49\textwidth}
            \includegraphics[trim={0 1.3cm 0 0},clip,width=\textwidth]{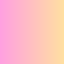}
            \caption{Range for \textit{horizon}.}
            \label{fig:randcol:sunset}
        \end{subfigure}
        \hfill
        \begin{subfigure}{.49\textwidth}
            \includegraphics[trim={0 1.3cm 0 0},clip,width=\textwidth]{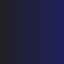}
            \caption{Range for \textit{night}.}
            \label{fig:randcol:night}
        \end{subfigure}
    \end{subfigure}
    \caption{Color ranges used to sample \textit{new white points} when applying \textit{horizon} and \textit{night} corruptions.}\label{fig:randcol}
\end{figure}

\item The \textit{horizon} corruption shifts the white-point of the input image to mimic haze on sunset/sunrise scenes. The new white points (RGB) are sampled uniformly in the range $[255, 192+\delta, 192-\delta]$, $\delta \sim  \mathcal{U}[-32, 32]$, as shown in Fig.~\ref{fig:randcol:sunset}.
An example is reported in the second row of Fig. \ref{fig:corrupions}.  

\item The \textit{night} corruption  simulates nighttime acquisition. We tackle this problem in two ways: (i) changing the white point to a dark blue (uniformly sampled in the color range $[32, 32, 64+\delta]$, $\delta \sim \mathcal{U}[-24, 24]$, as shown in Fig.~\ref{fig:randcol:night}) and (ii) darkening the brightest pixels of the scene (which tend to be found in the sky region of an outdoor image). The effect of this procedure is shown in the third row of Fig.~\ref{fig:corrupions}. 
 
\item The \textit{white-point} corruption %
mimics different behaviors of cameras' white-point balancing procedure, which could lead to images with unrealistic colors even if their relative chromaticity is consistent. %
In this case, %
we select a random white point using a uniform distribution $\mathcal{U}[223,287]$ (\ie, a variation of an eighth) for each of the 3 RGB components and then re-balance the image pixel values accordingly. A visual example is reported in the last row of Fig.~\ref{fig:corrupions}.
\end{enumerate}

\subsection{Inference Process}\label{sub:inference}
The goal of our system is to be as easy to use and modular as possible. Therefore, we strove to keep the inference process footprint minimal.
In Fig.~\ref{fig:abstract}, we reported a comparison between standard inference practice on a downstream task (top half), and the modified pipeline using our modules (bottom half).
Notice how our modules can be seamlessly embedded in any architecture to improve its final accuracy.

In detail, to use our system on a given input sample, we first normalize and standardize it (as detailed in Sec. \ref{sub:impl}).
Then, for faster processing, we downsize a copy of the input image to have the smallest dimension equal to $232$ pixels and extract a central square crop of $224$ pixels from it. This low-resolution cropped version is fed to the NEM, which predicts the parameters to enhance the image.
Finally, the predicted parameters and the full-resolution image are fed to the DWM to clean the image and enhance its content. 

The resulting output can either be used as-is, if the final objective was attaining a cleaned image, or fed to any downstream model to attain a more accurate prediction on the considered task.
Overall, the additional computational overhead is minimal, with our modules supporting up to about 300fps at full HD resolution (see  Sec.~\ref{subsub:cost}).

\begin{table}[!t]
    \centering
    \begin{tabular}{c|c|c}
        \textbf{Layer} & \textbf{Operations} & \textbf{Feature Size} \\
        \hline
        in & conv 3x3, 64 & 224x224x64 \\
        \hline
        \multirow{2}{*}{block1} & avg. pool, 5x5, stride 4 & \multirow{2}{*}{56x56x128} \\
        & 2 x ( conv 3x3, 128 ) \\
        \hline
        \multirow{2}{*}{block2} & avg. pool, 5x5, stride 4 & \multirow{2}{*}{14x14x256} \\
        & 2 x ( conv 3x3, 256 ) \\
        \hline
        \multirow{2}{*}{block3} & avg. pool, 5x5, stride 4 & \multirow{2}{*}{4x4x512} \\
        & 2 x ( conv 3x3, 512 ) \\
        \hline
        \multirow{2}{*}{out} & global avg. pool & \multirow{2}{*}{37} \\
        & fcn, 37 
    \end{tabular}
    \caption{Breakdown of our NEM architecture. BatchNorm layers and ReLU activations are added after each convolution.} 
    \label{tab:NEM}
\end{table}

\section{Results and Discussion}
To validate the generalization capability of our modular system, we evaluate its performance on two main tasks: image classification (Sec. \ref{sec:res:clas}),  and semantic segmentation (Sec. \ref{sec:res:seg}).
In the following, we will report the quantitative and qualitative results attained in various tasks by discussing them and drawing comparisons with competing strategies. 

\subsection{Implementation Details}\label{sub:impl}

\begin{table*}[t]
    \centering
    \resizebox{\textwidth}{!}{%
    \setlength{\tabcolsep}{2.5pt}
    \begin{tabular}{ccc|cccc:cccc:ccc:cccc|c:c|c:c}
    & & & \multicolumn{4}{|c:}{Blur} & \multicolumn{4}{:c:}{Digital} & \multicolumn{3}{:c:}{Noise} & \multicolumn{4}{:c|}{Weather} \\
    \cdashline{4-18}
    Arch. & Pre-train & Ours & Defocus & Glass & Motion & Zoom & Contrast & Elastic & JPEG & Pixelate & Gaussian & Impulse & Shot & Brightness & Fog & Frost & Snow & Clean & Corr.\ Avg. & $\Delta$ & $\%\Delta$ \\
    \hline
    \multirow{10}{*}{RN50} & \multirow{2}{*}{V2~\cite{torchvision2016}} & \xmark & 43.6 & 29.7 & 43.2 & 42.9 & 60.0 & 48.8 & 61.4 & 53.0 & 42.2 & 37.2 & 40.5 & 72.6 & 59.5 & 47.0 & 41.2 & 80.7 & 48.2 & - & - \\
    &  & \cmark & 44.1 & 36.0 & 61.2 & 46.8 & 61.8 & 51.4 & 61.7 & 61.1 & 43.6 & 39.0 & 42.0 & 72.0 & 62.1 & 49.7 & 51.0 & 80.5 & 52.2 & 4.0 & 8.3 \\
    \cdashline{2-22}
    &  \multirow{2}{*}{V1~\cite{torchvision2016}} & \xmark & 37.8 & 27.7 & 38.4 & 35.9 & 37.1 & 44.2 & 55.1 & 49.0 & 35.4 & 31.3 & 33.2 & 66.1 & 43.4 & 36.8 & 30.7 & 75.7 & 40.1 & - & -\\
    &  & \cmark & 36.1 & 30.6 & 52.5 & 38.2 & 33.4 & 45.0 & 55.4 & 56.5 & 35.9 & 32.0 & 33.7 & 64.4 & 38.4 & 38.7 & 40.4 & 75.4 & 42.1 & 2.0 & 5.0 \\
    \cdashline{2-22}
    & \multirow{2}{*}{HA~\cite{yucel2023hybridaugment++}} & \xmark & 60.0 & 47.8 & 60.3 & 48.6 & 55.2 & 52.9 & 58.9 & 70.3 & 61.2 & 61.3 & 61.2 & 70.9 & 56.6 & 54.4 & 49.6 & 75.1 & 57.9 & - & - \\
    &  & \cmark & 60.2 & 50.9 & 68.9 & 50.0 & 55.5 & 54.5 & 60.3 & 71.3 & 61.7 & 61.4 & 61.7 & 70.7 & 54.8 & 58.0 & 58.1 & 75.1 & 59.9 & 2.0 & 3.5 \\
    \cdashline{2-22}
    & \multirow{2}{*}{PRIME~\cite{modas2022prime}} & \xmark & 46.9 & 38.9 & 47.4 & 45.3 & 56.6 & 56.1 & 61.9 & 60.5 & 55.2 & 55.2 & 55.3 & 70.1 & 52.3 & 48.9 & 44.2 & 76.9 & 53.0 & - & - \\
    & & \cmark & 46.6 & 42.7 & 61.9 & 47.0 & 58.8 & 56.5 & 61.9 & 64.3 & 56.4 & 56.1 & 56.1 & 69.2 & 54.1 & 50.0 & 50.3 & 76.5 & 55.5 & 2.5 & 4.7 \\
    \cdashline{2-22}
    & \multirow{2}{*}{PIXMIX~\cite{Hendrycks2021PixMixDP}} & \xmark & 43.9 & 30.5 & 43.3 & 40.8 & 59.1 & 47.1 & 61.5 & 54.6 & 47.9 & 46.4 & 46.4 & 70.4 & 55.9 & 49.6 & 47.9 & 77.8 & 49.7 & - & -\\
    &  & \cmark & 44.1 & 35.5 & 60.7 & 44.5 & 60.4 & 49.1 & 61.5 & 59.3 & 49.6 & 48.0 & 48.4 & 69.6 & 56.8 & 51.0 & 52.5 & 77.7 & 52.7 & 3.0 & 6.0 \\
    \midrule
    \multirow{2}{*}{VGG16} & \multirow{2}{*}{V1~\cite{torchvision2016}} & \xmark & 25.2 & 20.0 & 28.0 & 28.3 & 28.8 & 35.7 & 43.0 & 34.0 & 21.9 & 17.7 & 19.8 & 58.9 & 35.2 & 26.8 & 22.4 & 71.4 & 29.7 & - & - \\
    & & \cmark & 22.3 & 21.6 & 39.5 & 28.6 & 24.4 & 36.2 & 43.4 & 46.0 & 23.0 & 19.8 & 21.2 & 56.9 & 28.2 & 29.2 & 31.2 & 71.2 & 31.4 & 1.7 & 5.7 \\
    \hline
    \multirow{2}{*}{Swin-T} & \multirow{2}{*}{V1~\cite{torchvision2016}} & \xmark & 45.2 & 35.5 & 52.4 & 42.4 & 64.1 & 53.2 & 64.1 & 58.2 & 52.2 & 48.9 & 49.0 & 73.5 & 61.8 & 56.7 & 50.9 & 81.3 & 53.9 & - & -\\
    &  & \cmark & 44.9 & 38.6 & 62.1 & 43.6 & 63.7 & 55.1 & 63.9 & 63.8 & 52.8 & 49.5 & 49.6 & 72.8 & 62.3 & 57.2 & 56.4 & 80.7 & 55.7 & 1.8 & 3.3 \\
    \hline
    \multirow{2}{*}{CLIP} & \multirow{2}{*}{V1~\cite{radford2021learning}} & \xmark & 22.4 & 14.4 & 21.9 & 21.1 & 28.4 & 25.6 & 34.5 & 29.7 & 18.3 & 12.9 & 16.7 & 45.2 & 33.4 & 23.4 & 18.9 & 55.2 & 24.5 & - & -\\
    & & \cmark & 18.6 & 17.3 & 30.8 & 21.4 & 25.4 & 26.1 & 34.9 & 35.4 & 18.4 & 12.9 & 17.2 & 43.7 & 29.9 & 24.2 & 24.3 & 55.1 & 25.4 & 0.9 & 3.7
    \end{tabular}}
    \caption{Results for the image classification task on the ImageNetC dataset (higher is better).}
    \label{tab:inc}
\end{table*}

\textbf{Experimental Setup.}
To highlight the generalization capability of our approach, we trained our modules only once on the upstream classification task on the ImageNet dataset. 
The optimization lasted for $50k$ steps, using batch size $384$ and Adam~\cite{kingma2014adam} optimizer with learning rate ${\small 10}^{-3}$ scheduled according to a cosine annealing strategy.
We remark that when optimizing \mname[], the upstream module can be completely frozen, reducing the computational complexity of training. This is the scenario we consider in our work. Furthermore, the training is done once and the same pre-trained weights are used for all the downstream models and tasks.

Our NEM is implemented using a 3-block architecture with strong downsampling (\ie, with stride $4$) between layers. A detailed breakdown is reported in Tab.~\ref{tab:NEM}.
In total, our module uses seven 2D convolutions and a single fully connected layer to project the downsampled features into the space of parameters $(\textbf{K}, \mathbf{C}_{M}, \mathbf{C}_{S}) \in \mathbb{R}^{A}$. We considered a kernel size $K=5$, which yields $A \coloneqq d(d+1) + K^2 = 37$ in our setup.
During training, the exponential moving average rate was defined as $\beta = 0.9$, and the weight factor for the regularization loss was set to $\lambda_{EMA} = 0.5$.
Without loss of generality, the normalization of the input images at inference-time must match the normalization seen during training.
In our case, $\mathbf{X}$ is normalized in the range $[0,1]$ and standardized using mean $\mu = [0.485, 0.456, 0.406]$ and standard deviation $\sigma = [0.229, 0.224, 0.225]$.

\textbf{Datasets.}
We employed our system on several real and synthetic datasets to verify its effectiveness
in different scenarios for the image classification task (Sec.~\ref{sec:res:clas}). In particular, we used the ImageNetC~\cite{hendrycks2019benchmarking} and ImageNetC-Bar~\cite{mintun2021interaction} datasets having synthetic corruptions (both based on the ImageNet~\cite{russakovsky2015imagenet}) and the real dataset VizWiz~\cite{bafghi2023new} having natural corruptions.
We also introduced ImageNetC-mixed\footnote{The split will be released for reproducibility.}, a new benchmark to investigate the reliability of models when presented with multiple corruptions at once. We built it by randomly applying 1-to-3 corruptions to the same image, sampled from the joint pool of augmentations of our data augmentation pipeline and those proposed in ImageNetC. 

For what concerns semantic segmentation (Sec.~\ref{sec:res:seg}), we analyzed the models using driving scenes tackling the domain adaptation problem for clear-to-adverse weather conditions.
In this case, we employed three real world datasets: i) Cityscapes~\cite{Cordts2016Cityscapes} as the training  (source) domain; ii) ACDC~\cite{SDV21} and iii) DarkZurich~\cite{sakaridis2019guided} as testing downstream (target) domains.

\textbf{Metrics.} In image classification, we report the per-corruption accuracies (\textit{Acc}, $\uparrow$), their mean (\textit{Corr. Avg.}, $\uparrow$), and the the accuracy on corruption-free data (\textit{Clean}, $\uparrow$). 
In semantic segmentation, we report the results as per-class IoU (Intersection over Union) or its mean (mIoU, $\uparrow$).
In all cases, $\Delta$ ($\%\Delta$) refers to the absolute (relative) gain with respect to the considered baseline. 

\subsection{Results for Image Classification}
\label{sec:res:clas}
In the image classification task, we investigate three main scenarios: i) single synthetic corruptions (using ImageNetC~\cite{hendrycks2019benchmarking}) in Sec.~\ref{subsub:inc}, ii) multiple synthetic corruptions (using
ImageNetC-mixed) in Sec.~\ref{subsub:mixed}, and iii) real-world corruptions (using VizWiz~\cite{bafghi2023new}) in Sec.~\ref{subsub:vizwiz}. 
Finally, we analyze the computational cost in Sec.~\ref{subsub:cost}.

\subsubsection{\textbf{Results on Single Synthetic Corruptions (ImageNetC)}}\label{subsub:inc}
For this discussion, we refer to Table~\ref{tab:inc}, where we report the final accuracy attained by our approach when mounted on multiple different backbones. 
In particular, we use a ResNet50~\cite{he2015deep} with different pre-training weights (TorchVision~\cite{torchvision2016} V2 and V1, HA~\cite{yucel2023hybridaugment++}, PRIME~\cite{modas2022prime}, and PIXMIX~\cite{Hendrycks2021PixMixDP}), a VGG16~\cite{simonyan2014very}, a Swin-Tiny~\cite{liu2021swin}, and finally the CLIP~\cite{radford2021learning} foundation model.
Notably, we train our \mname model only once using ResNet50 with TorchVision's V2 weights, and we use the trained model as-is on all other experiments, highlighting the generalization capability of our approach. 

Our \mname consistently improves the average accuracy across the considered models with an average gain of $2.2\%$ in absolute terms and an average relative gain of $5.0\%$. 
As anticipated, a remarkable feature of \mname is the capability to improve the performance even for architectures designed to work well in the ImageNetC task, such as HA~\cite{yucel2023hybridaugment++}, PRIME~\cite{modas2022prime}, and PIXMIX~\cite{Hendrycks2021PixMixDP}. 
These are three strong  state-of-the-art data augmentation approaches and, therefore, can handle corrupted samples from ImageNetC better than weaker data augmentation approaches. 
Keeping this in mind, the improvement brought by our approach when used jointly with these architectures is even more striking, signifying that the effect of our module is complementary to existing state-of-the-art approaches, and using both strategies together can improve the absolute accuracy significantly. 
Numerically, when using our system together with HA, we are able to improve the performance over TorchVision's V1 baseline by almost $20\%$ in absolute terms ($49.4\%$ relative). 

Moreover, the gain is well-spread across the various corruptions and, in cases particularly suited to be tackled by our constrained model, we can appreciate some higher gains. For example, in \textit{Motion Blur} and \textit{Snow}, we get significant performance gains (up to $9\%$ absolute points) even on already-robust backbones like those pre-trained via HA. 
We believe that the \textit{motion blur} kernel can be estimated accurately with our learned filter, whereas our affine color transforms handle the low contrast brought by \textit{snow}, regardless of the localized bright spots. We experience slight performance drops on the \textit{fog} corruption, due to changes to the frequency distribution our system may introduce via the global filter. Methods such as HA are designed with frequency-spectra changes in mind. Therefore, they expect a distribution of input images with specific frequency characteristics.
Another performance drop is observed on the \textit{Brightness} corruption. This corruption is modeled as a non-linear change in the HSV color space, and our affine color transforms could not approximate the underlying function accurately. 
Note that these limitations introduce only marginal drops in accuracy, and do not change the overall improvements brought by our system.

\begin{table}[t]
    \centering
    \begin{tabular}{c|cc|cc}
        Pre-training & Base & Ours & $\Delta$ & $\%\Delta$ \\
        \hline
        V2~\cite{torchvision2016} & 38.2 & \textbf{41.0} & 2.8 & 7.3 \\
        V1~\cite{torchvision2016} & 31.2 & \textbf{32.9} & 1.7 & 5.5\\
        HA~\cite{yucel2023hybridaugment++} & 47.2 & \textbf{48.9} & 1.7 & 3.6\\
        PRIME~\cite{modas2022prime} & 43.4 & \textbf{44.9} & 1.5 & 3.5\\
        PIXMIX~\cite{Hendrycks2021PixMixDP} & 41.1 & \textbf{43.2} & 2.1 & 5.1\\
    \end{tabular}
    \caption{Accuracy on ImageNetC-mixed via ResNet50.}    \label{tab:inc_multi}
\end{table}

\begin{table}[t]
    \centering
    \setlength{\tabcolsep}{3pt}
    \resizebox{\linewidth}{!}{%
    \begin{tabular}{c|cccccc|c:c}
         Ours & Blur & Bright & Frame & Rotation & Obscured & Dark & Clean & Corr.\ Avg. \\
        \hline
         \xmark & 44.4 & \textbf{40.9} & \textbf{42.7} & 38.5 & 37.9 & 44.3 & 48.6 & 41.5 \\
         \cmark & \textbf{44.6} & \textbf{40.9} & 42.6 & \textbf{39.2} & \textbf{39.7} & \textbf{45.3} & \textbf{48.8} & \textbf{42.0} \\ 
    \end{tabular}}
    \caption{Results for the VizWiz with ResNet50 V2.}
    \label{tab:vizwiz}
\end{table}

Lastly, we discuss the results when employing our approach with downstream architectures that are different than the one used for pre-training our modules.
We verify the usefulness of our method on the widely used  VGG16 convolutional architecture and on a transformer-based architecture (\eg, Swin-Tiny). 
Then, we considered the CLIP foundation model~\cite{radford2021learning} that was pre-trained to align image-text embeddings to the same semantic value, rather than explicitly recognizing the input image category.
Our modules bring improvement even in this case, highlighting how the approach is able not only to change the graphical appearance but also to highlight the semantic content, making it easier for the downstream network.

\subsubsection{\textbf{Results on Mixed Synthetic Corruptions (ImageNetC-mixed)}}\label{subsub:mixed}
Following the tests on a single corruption at a time, we consider the more challenging setting of multiple corruptions together using the hereby proposed ImageNetC-mixed dataset. (refer to the datasets section in \ref{sub:impl} for more information).
The results of these analyses are shown in Table~\ref{tab:inc_multi}.
Despite the existing data augmentation approaches losing about 10\% accuracy compared with the single corruption case, our approach maintains a stable gain of $2.0\%$ in absolute terms, corresponding to $5.0\%$ relative improvement. This proves the ability of our approach to handle the composition of input corruptions, as it is often encountered in practice.

\subsubsection{\textbf{Results on Real-World Corruptions (VizWiz)}}\label{subsub:vizwiz}
In the previous sections, we have investigated the performance of our modules on synthetic corruptions, following the standard evaluation pipeline \cite{hendrycks2019benchmarking,hendrycks2020augmix}. 
However, it is important to verify that the performance improvement is maintained even on real data. To this end, we used the corrupted real-world VizWiz dataset~\cite{bafghi2023new}. 
The numerical results are reported in Table~\ref{tab:vizwiz}, where we observe a  $1.2\%$ relative gain on corrupted images, while also improving the accuracy on clean images by relative $0.4\%$.

\subsubsection{\textbf{Computational Cost}}\label{subsub:cost}

We report the complexity of our approach in FLOPs (Floating Point Operations) %
in Table~\ref{tab:complexity}, comparing it to other architectures that can be used as (or converted into) input-level denoisers.
We computed the cost of our approach using the PTFlops library~\cite{ptflops} at a resolution of $1920 \times 1080 \text{px}^2$.

Our module is one to four orders of magnitude faster than the diffusion model competitors.
Moreover, its computational complexity does not scale significantly with the input resolution - since a resizing stage is done before parameter estimation and the only operations applied on the full-resolution image are the spatial convolution and a matrix multiplication over the channels (see Sec.~\ref{sub:inference}). This means that we can handle images of arbitrary resolution, contrary to what happens for, \eg, the fixed-size diffusion models.
Similar considerations also hold for auto-encoders and GANs.
This confirms the efficiency of our strategy.

Finally, we computed the throughput of our module on an NVIDIA GTX 1080Ti GPU, obtaining a speed of 289.4 images/second; corresponding to a per-image inference time of about 3ms, meaning that our approach can be easily employed in real-time applications with minimal impact on the inference time of downstream architectures.

\begin{table}[t]
    \centering
    \begin{tabular}{cc|c}
        & Model & GFLOPs \\
        \hline
        & \mname (ours) & 2.0 \\
        \hdashline
        \multirow{6}{*}{\rotatebox{90}{Diffusion}} & LDM-8~\cite{ulhaq2022efficient} & 37.1 \\
        & Frido~\cite{ulhaq2022efficient} & 37.3 \\
        & Dido-gating~\cite{ulhaq2022efficient} & 39.7 \\
        & Diff 64~\cite{dhariwal2021diffusion} & 658.0 \\
        & Diff 128~\cite{dhariwal2021diffusion} & 1860.0 \\
        & Diff 256~\cite{dhariwal2021diffusion} & 6700.0 \\
        \hdashline
        \multirow{6}{*}{\rotatebox{90}{Auto-Encoder}} & UNet~\cite{akbari2020generalized} & 103.0 \\
        & GoConv-UNet~\cite{akbari2020generalized} & 67.7 \\
        & OrgOct-UNet~\cite{akbari2020generalized} & 65.6 \\
        & MSUDCAE~\cite{zhang2023color} & 37.5 \\
        & MSDCAE~\cite{zhang2023color} & 23.5 \\
        & CBAM\_MSUDCAE~\cite{zhang2023color} & 37.4 \\
        \hdashline
        \multirow{6}{*}{\rotatebox{90}{GANs}} & CycleGAN~\cite{wang2020gan} & 52.9 \\
        & U-GAT-IT-light~\cite{chen2020reusing} & 105.0 \\
        & NICE-GAN~\cite{chen2020reusing} & 67.6 \\
        & N2D-GAN~\cite{li2022spn2d} & 43.6 \\
        & SPN2D-GAN~\cite{li2022spn2d} & 94.0 \\
        & STGAN~\cite{nie2021gigan} & 519.0 \\
    \end{tabular}
    \caption{Computational complexity of our method compared to other input-level image enhancement strategies.
    }
    \label{tab:complexity}
\end{table}

{
\begin{table*}[t]
    \centering
    \resizebox{\textwidth}{!}{%
    \begin{tabular}{cc|ccccccccccccccccccc|c|c:c}
        Model & Dataset & \rotatebox{90}{Road} & \rotatebox{90}{Sidewalk} & \rotatebox{90}{Building} & \rotatebox{90}{Wall} & \rotatebox{90}{Fence} & \rotatebox{90}{Pole} & \rotatebox{90}{T. Light} & \rotatebox{90}{T. Sign} & \rotatebox{90}{Vegetation} & \rotatebox{90}{Terrain} & \rotatebox{90}{Sky} & \rotatebox{90}{Person} & \rotatebox{90}{Rider} & \rotatebox{90}{Car} & \rotatebox{90}{Truck} & \rotatebox{90}{Bus} & \rotatebox{90}{Train} & \rotatebox{90}{Motorcycle} & \rotatebox{90}{Bicycle} & mIoU & $\Delta$ & $\%\Delta$ \\
        \hline
        \multirow{2}{*}{Source Only} & ACDC & 73.5 & 33.2 & 62.1 & 21.3 & 20.6 & 23.3 & 43.4 & 34.1 & 69.6 & 25.1 & 75.4 & 38.9 & 15.9 & 65.9 & 29.7 & 30.2 & 36.6 & 16.5 & 23.0 & 38.9 & - & - \\
        & DarkZurich & 63.2 & 12.0 & 49.9 & 9.1 & 12.2 & 9.7 & 15.1 & 7.7 & 40.3 & 12.8 & 2.5 & 16.0 & 7.9 & 31.1 & - & - & 0.0 & 8.4 & 17.0 & 18.6 & - & -\\
        \hline
        \multirow{2}{*}{Ours} & ACDC & 68.6 & 36.2 & 60.1 & 22.7 & 22.2 & 25.5 & 46.0 & 35.6 & 72.3 & 25.9 & 72.4 & 42.7 & 18.3 & 68.4 & 35.0 & 30.2 & 39.5 & 19.6 & 27.7 & 40.5 & 1.6 & 4.1\\
        & DarkZurich & 53.0 & 15.9 & 46.6 & 10.7 & 13.3 & 10.7 & 21.4 & 8.4 & 45.7 & 12.1 & 1.0 & 16.8 & 7.8 & 34.2 & - & - & 0.0 & 10.0 & 21.0 & 19.3 & 0.7 & 3.8 \\
        \hline
    \end{tabular}}
    \caption{Quantitative results for semantic segmentation using a DeepLabV2~\cite{chen2017deeplab} architecture with the ResNet50 backbone.}
    \label{tab:City2Adverse}
\end{table*}
\begin{figure*}
    \begin{subfigure}{\textwidth}
        \centering
        \begin{subfigure}{.19\textwidth}
            \includegraphics[width=\textwidth]{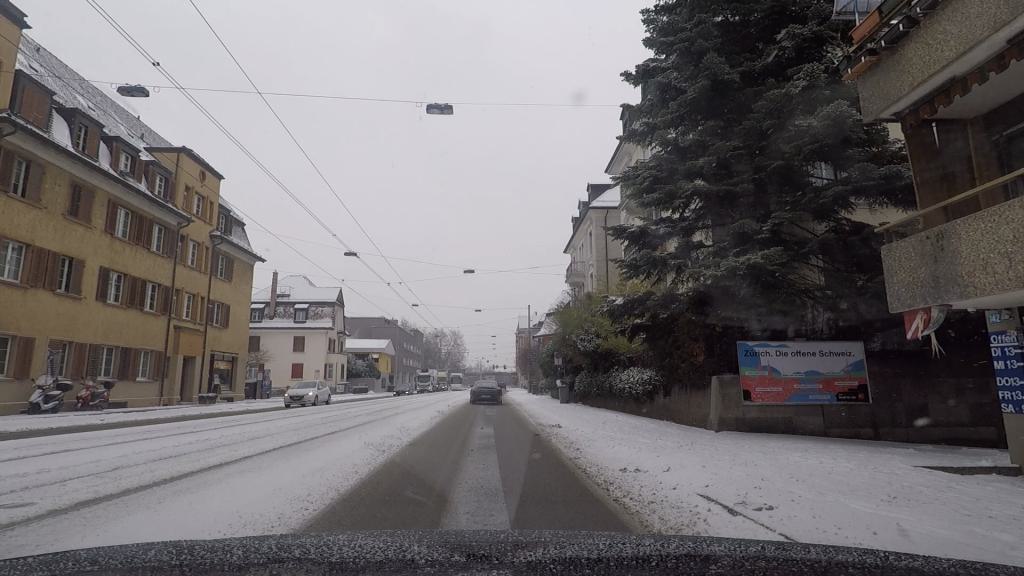}
        \end{subfigure}
        \begin{subfigure}{.19\textwidth}
            \includegraphics[width=\textwidth]{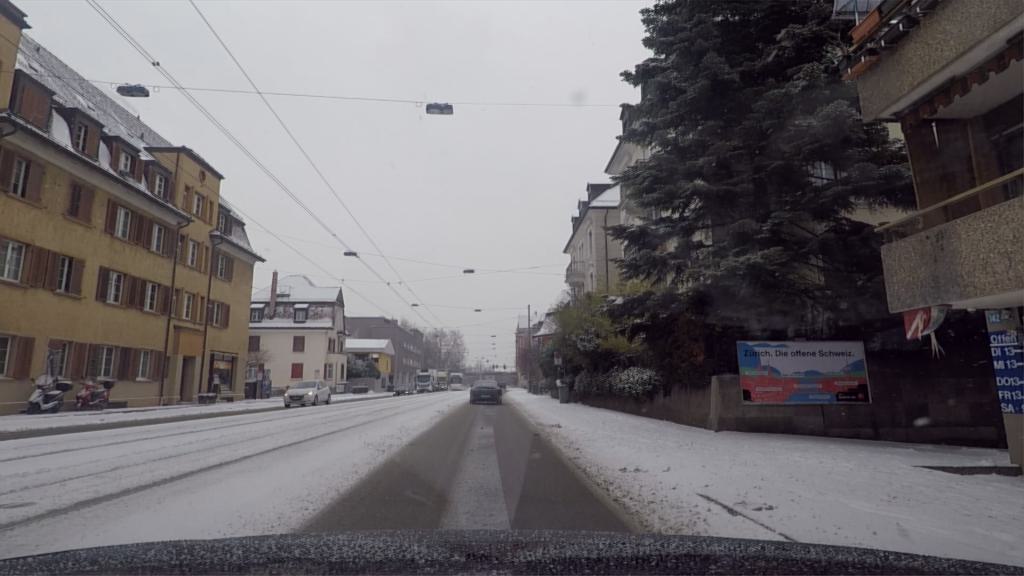}
        \end{subfigure}
        \begin{subfigure}{.19\textwidth}
            \includegraphics[width=\textwidth]{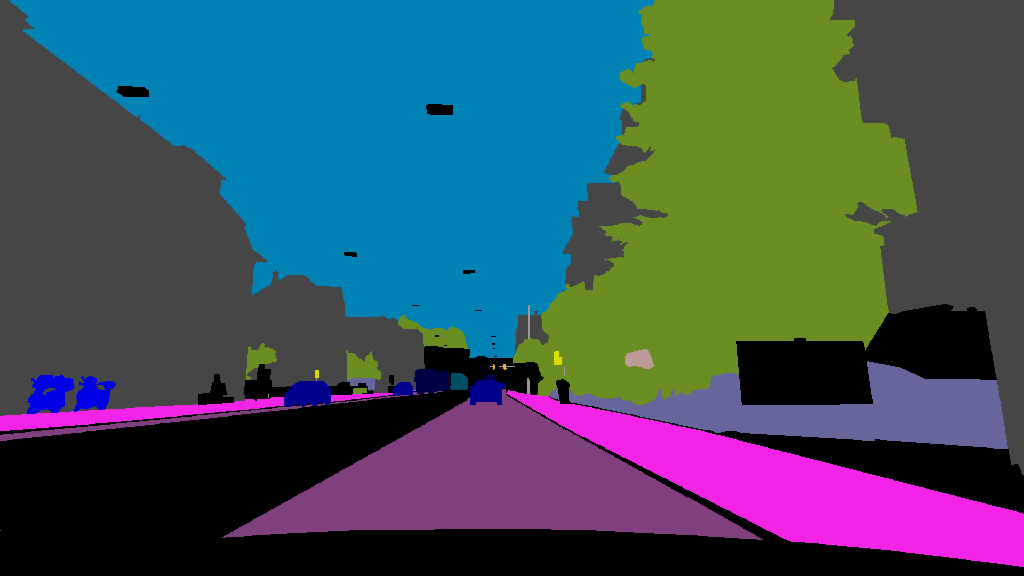}
        \end{subfigure}
        \begin{subfigure}{.19\textwidth}
            \includegraphics[width=\textwidth]{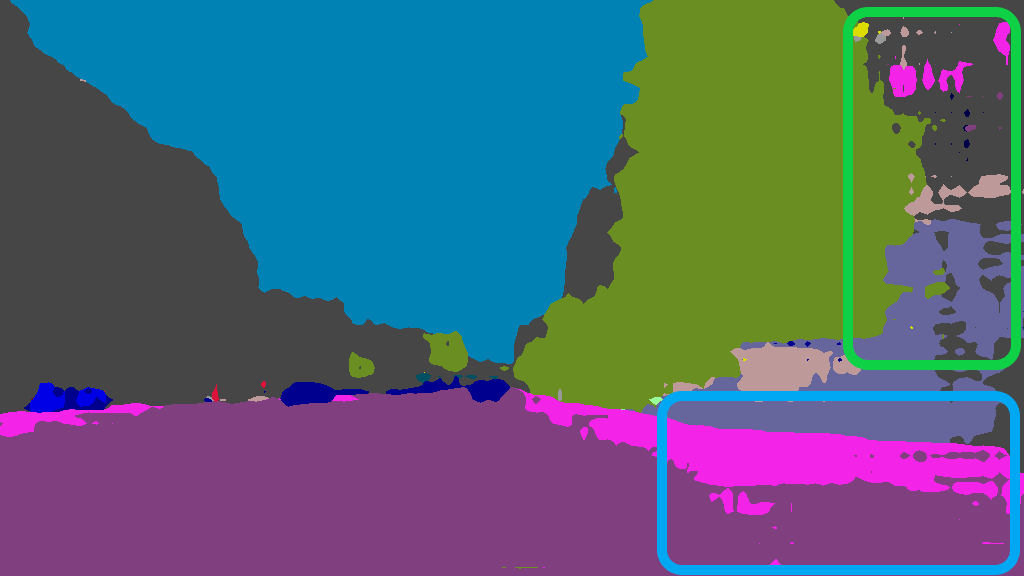}
        \end{subfigure}
        \begin{subfigure}{.19\textwidth}
            \includegraphics[width=\textwidth]{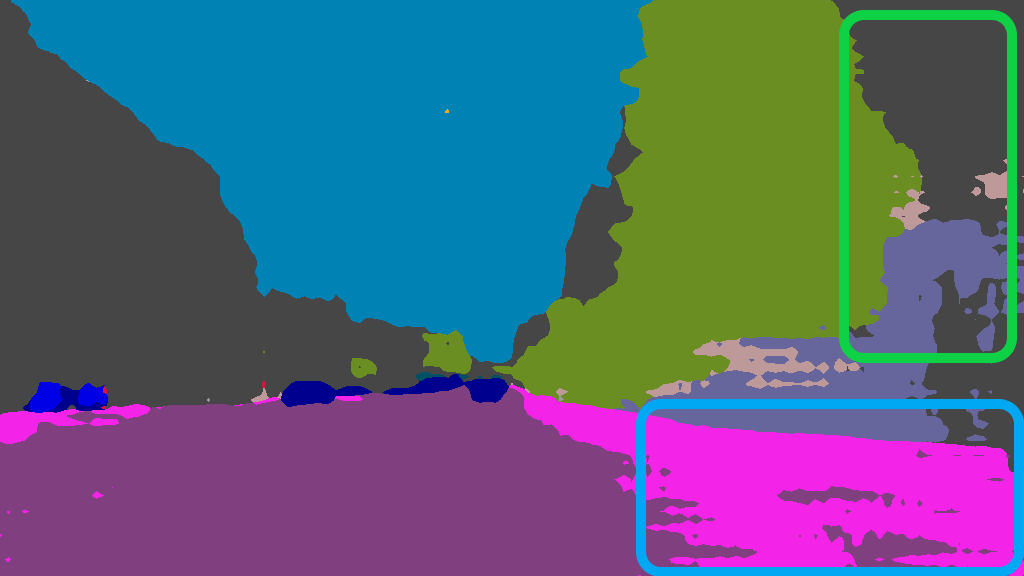}
        \end{subfigure}
    \end{subfigure}
    \begin{subfigure}{\textwidth}
        \centering
        \begin{subfigure}{.19\textwidth}
            \includegraphics[width=\textwidth]{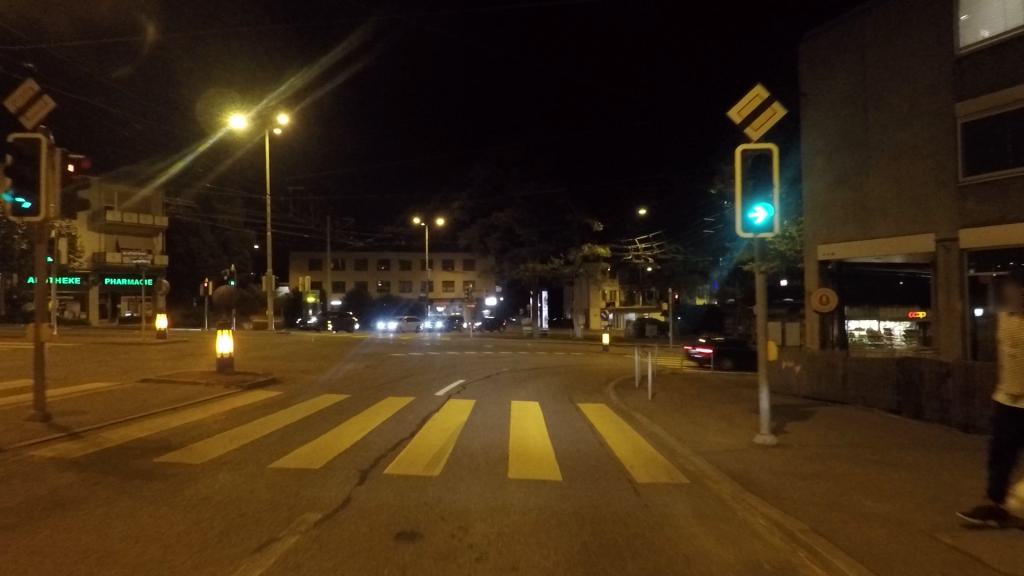}
            \caption*{Original}
        \end{subfigure}
        \begin{subfigure}{.19\textwidth}
            \includegraphics[width=\textwidth]{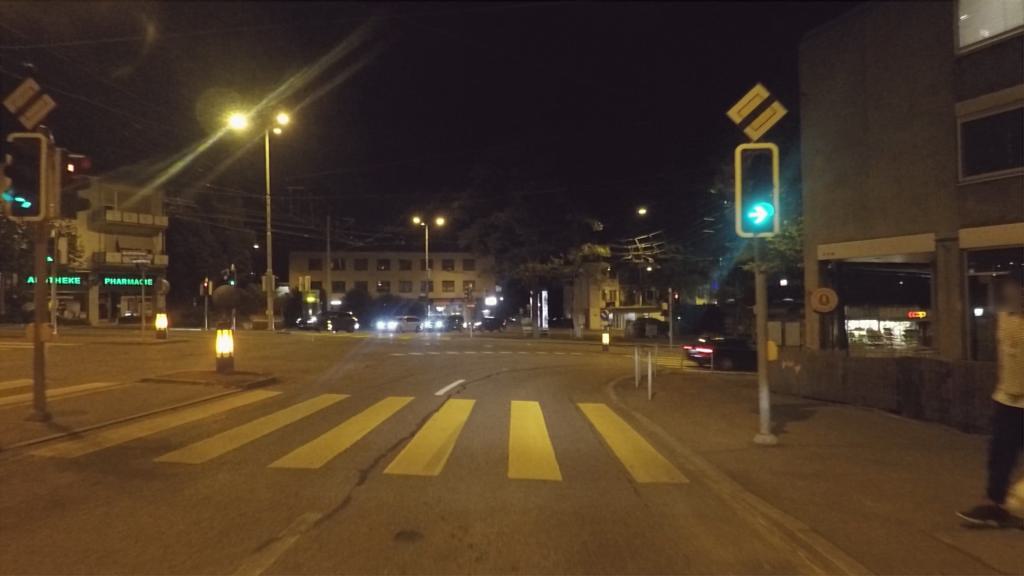}
            \caption*{Warped}
        \end{subfigure}
        \begin{subfigure}{.19\textwidth}
            \includegraphics[width=\textwidth]{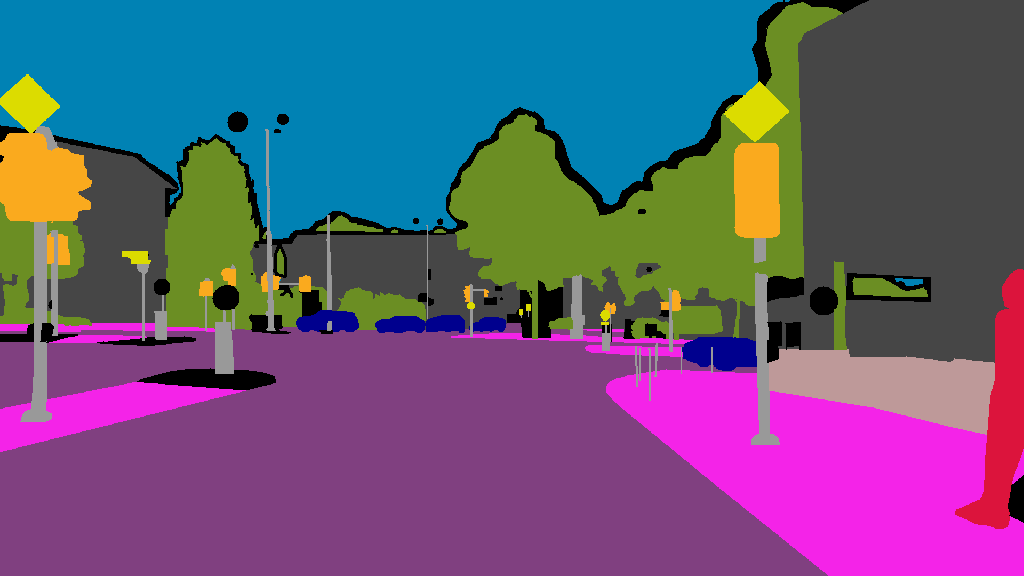}
            \caption*{Ground Truth}
        \end{subfigure}
        \begin{subfigure}{.19\textwidth}
            \includegraphics[width=\textwidth]{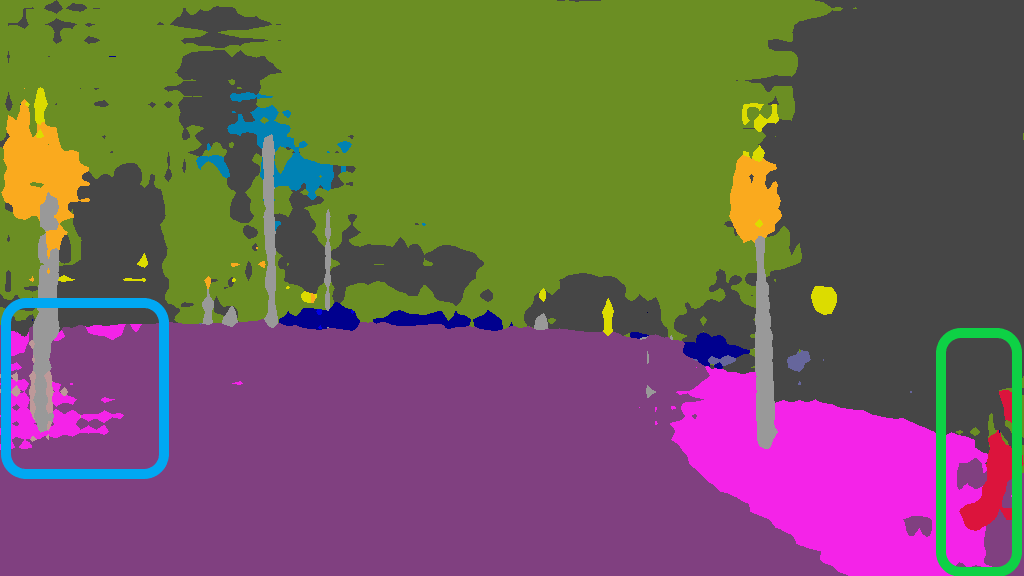}
            \caption*{Original Prediction}
        \end{subfigure}
        \begin{subfigure}{.19\textwidth}
            \includegraphics[width=\textwidth]{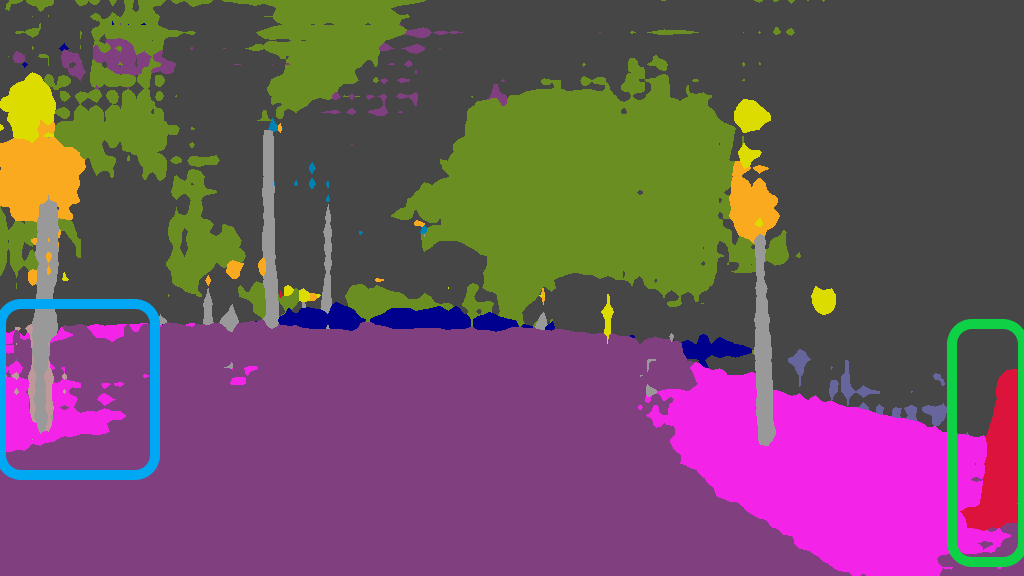}
            \caption*{Warped Prediction}
        \end{subfigure}
    \end{subfigure}
    \caption{Qualitative results on the ACDC semantic segmentation benchmark.}
    \label{fig:quali_acdc}
\end{figure*}
}

\begin{table*}[t]
    \centering
    \resizebox{\linewidth}{!}{%
    \setlength{\tabcolsep}{4pt}
    \begin{tabular}{cc|cccccccccc|c:c|c:c}
    Pre-training & Ours & Blue & Brown & Caustic & Checker & C. Sine & Inv. Sparkle & Perlin & Plasma & S. Sine & Sparkle & Clean & Corr. Avg. & $\Delta$ & $\%\Delta$ \\
    \hline
    \multirow{2}{*}{V2~\cite{torchvision2016}} & \xmark & 61.1 & 66.5 & 59.3 & 57.2 & 19.3 & 28.7 & 67.4 & 36.4 & 30.0 & 69.0 & 80.7 & 48.5 & - & - \\
    & \cmark & 52.2 & 65.8 & 58.5 & 56.1 & 20.1 & 29.4 & 66.7 & 37.1 & 33.3 & 68.2 & 80.5 & 48.7 & 0.2 & 0.4\\
    \hline
    \multirow{2}{*}{HA~\cite{yucel2023hybridaugment++}} & \xmark & 64.3 & 64.0 & 58.7 & 48.7 & 40.1 & 25.1 & 68.8 & 31.5 & 57.4 & 64.8 & 75.1 & 52.3 & - & -\\
    & \cmark & 64.7 & 63.5 & 58.6 & 48.0 & 41.1 & 26.9 & 68.9 & 31.6 & 59.1 & 64.7 & 75.1 & 52.6 & 0.3 & 0.6\\
    \end{tabular}}
    \caption{Quantitative results on the ImageNetC-Bar dataset with ResNet50.}
    \label{tab:incb}
\end{table*}

\subsection{Results for Semantic Segmentation}
\label{sec:res:seg}

The analyses up to this point have been in the same task as training (\ie, image classification) with some changes in the downstream architecture. 
However, to truly show the generalization capabilities of our approach, we change the downstream task completely and verify that the accuracy improvement is preserved.
For this investigation, we choose the semantic segmentation task and, in particular, domain adaptation to adverse weather conditions. We train the baseline architecture used in \cite{barbato2021latent,barbato2021road} (DeepLabV2~\cite{chen2017deeplab} with ResNet50 backbone~\cite{resnet}) on the Cityscapes~\cite{Cordts2016Cityscapes} dataset, and deploy it on two adverse weather datasets, namely, ACDC~\cite{SDV21} and DarkZurich~\cite{sakaridis2019guided}.

The quantitative results of this study are reported in Table~\ref{tab:City2Adverse}, where our architecture improves the mIoU score by $4.1\%$ and $3.8\%$ relative gain on ACDC and DarkZurich, respectively.
We report some qualitative results on ACDC in Fig.~\ref{fig:quali_acdc}, showing the original RGB image, the image warped by our architecture, and the corresponding predictions made by the segmentation model. 
In the first image (\textit{snow} scenario), our model improves the overall contrast. This aids in distinguishing the sidewalk on the right of the scene (light blue box), which was missed almost completely by the segmentation system when shown the original data. Moreover, it helps to recognize more accurately the building (light green box).
In the second image (\textit{night} scene), instead, our method brightens the scene, allowing us to identify the person on the side (light green box), which was not recognized by the source only architecture. We are also able to slightly improve the precision of the sidewalk on the left (light blue box).

\subsection{Ablation Studies}
\label{sec:ablation}

We report some ablation studies, namely: an analysis of corruptions that cannot be modeled by our strategy (ImageNetC-Bar, Sec.~\ref{subsub:incb}); a study on the iterative application of our module, which is a common failure point for denoisers and image enhancers (Sec.~\ref{subsub:iter}); a proof of concept for  applicability on video compression (Sec.~\ref{sec:res:video}).

\begin{figure*}[t]
    \centering
    \includegraphics[width=\linewidth]{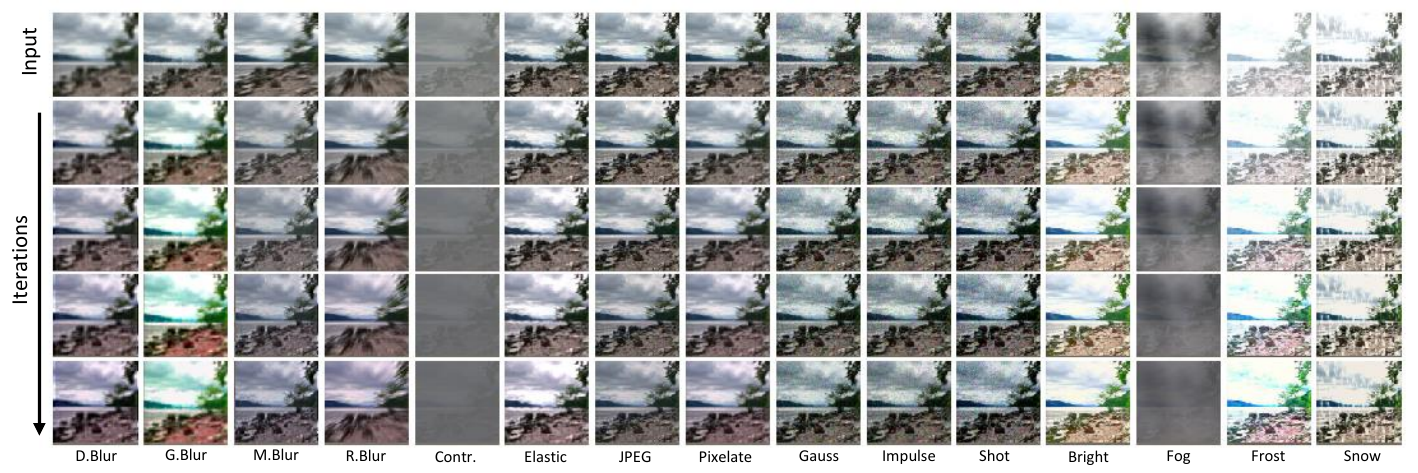}
    \caption{Qualitative results on iterated application of our method on an input image from the ImageNetC validation set.
    The first row represents the input, while the others correspond to four iterative applications of our modules.}
    \label{fig:res:iter}
\end{figure*}

\subsubsection{\textbf{Results on corruptions that cannot be modeled by \mname by construction (ImageNetC-Bar)}}\label{subsub:incb}

We further analyze the reliability of our method to corruptions that cannot be modeled by our modules. For this purpose, we test its performance on the ImageNetC-Bar dataset~\cite{mintun2021interaction}, which provides a corruption set complementary to ImageNetC. In general, the corruptions present in the dataset cannot be suitably modeled by our approach, since most of them contain non-spatially-uniform distortions. 

Nonetheless, we observe small accuracy gains when using our modules. This means that, even if we are not able to model the corruptions in the dataset, \mname can modify the inputs to make the classification task easier.
Looking at the results, we can identify three corruptions where our method brings consistent and noticeable gains: \textit{Inverse Sparkle}, \textit{Plasma}, and \textit{Single Sine}. 
In the first corruption, our method improves the accuracy by an average of $2.5\%$ in absolute terms ($4.8\%$ relative gain), in the second we improve by $0.4\%$ ($1.1\%$), while in the third we improve by $2.5\%$ ($7.0\%$). These three corruptions, in some way, are similar to those present in ImageNet-C and can be dealt with more effectively by our approach. In particular, the \textit{Plasma} noise is a colored version of \textit{Fog}; \textit{Inverse Sparkle} is similar to \textit{Brightness} in some instances, and the effect of \textit{Single Sine} is similar to \textit{Contrast} and \textit{Fog}. 

\subsubsection{\textbf{Iterative Application}}\label{subsub:iter}
Furthermore, we investigate a common fallacy of denoising and enhancing models, especially those based on autoencoder architectures~\cite{yacoby2020failure}: when fed their own outputs iteratively, their prediction may collapse into undesirable modes, or destroy completely the input content.
To evaluate qualitatively the susceptibility of our approach to this scenario, we fed the predictions provided by our system to itself four times and collected the results in Fig.~\ref{fig:res:iter}. 
The first row contains the image corrupted with the fifteen strategies provided by the dataset with a mid-severity level, while the other rows show the outputs of iterative applications of our approach. 

From the figure, we see that our architecture does not suffer from the mentioned problem, and all outputs are consistent with the original input. 
Moreover, in many cases, one can see a clear improvement following the subsequent application of our module. The most striking cases are the \textit{motion blur}, \textit{frost}, and \textit{snow} corruptions highlighting the potentiality for deployment of our architecture.
In the first case, our approach is able to completely remove the blurring artifacts, while in the other two, it removes the haze from the image and increases contrast.
Also, our method enhances significantly the overall image content and colors, while removing the blur at the same time for the \textit{Glass Blur} (second column).

\begin{table}[t]
    \centering
    \begin{tabular}{ccc|c:c}
        Compression Rate & MSE \xmark & MSE \cmark & $\Delta$ & $\%\Delta$ \\
        \hline
        45 & 0.087 & 0.085 & -0.002 & -2.6 \\ 
        50 & 0.117 & 0.113 & -0.005 & -4.0 \\ 
        55 & 0.117 & 0.112 & -0.005 & -4.0 \\ 
        60 & 0.117 & 0.113 & -0.004 & -3.8 
    \end{tabular}
    \caption{Results for feature extraction from compressed video data.} 
    \label{tab:video}
\end{table}

\subsubsection{\textbf{Video Compression}}\label{sec:res:video}
Finally, we provide a proof-of-concept investigation of our proposed system for live-streaming applications (\eg,  for scene understanding applications on broadcasted data).
We downloaded a freely available video from YouTube\footnote{\url{https://www.youtube.com/watch?v=nTGXmmBTmzo}, accessed on 26 Nov 2023} and we compressed it multiple times using the VLC Media Player software\footnote{\url{https://www.videolan.org/}, accessed on 26 Nov 2023} at different compression rates, before extracting the frames from the videos for their analysis. 

We take each frame obtained from the original, compressed, and processed videos to a VGG19~\cite{simonyan2014very} architecture, and compare the similarity of the features extracted by the model (via mean squared error, MSE, $\downarrow$).
This metric is representative of the overall reconstruction fidelity as well as the potential applicability to an end task, since the extracted features are strongly correlated to the final predictions, regardless of the end task employed.
The results given in Table~\ref{tab:video} show that our method brings a $4.0\%$ ($2.6\%$) relative improvement in the high- (low-) compression regime in the linear scale.

\section{Conclusions}

In this paper, we introduced a novel, modular, and efficient system that predicts explicit parameters for content enhancement of an input image targeting improved accuracy on downstream tasks.
The computational footprint of the modules is minimal, being more than 10x faster than competing approaches (2GFLOPs in total), and enjoys a throughput of about 300 images per second. A key feature of our approach is the capability of training without any paired (clean/corrupted) input samples, but rather learning automatically the most suitable transformations of an input image using the supervision on an upstream task. Even more remarkably, once the system has been trained on a given upstream task, it can generalize to arbitrary downstream tasks without fine-tuning.
To confirm the generalization claim, we validate our approach on three classification datasets and two semantic segmentation datasets, achieving noticeable improvement in all of them.

In future work, we plan to introduce new parameterizations and operators to capture a richer representation of the input and estimate more accurately the enhancing parameters (\eg, features of frequency components, or non-spatially-uniform operators), while maintaining high efficiency and full differentiability, which are fundamental for our approach.
Another avenue of research will target compressed video data, investigating the effect of re-using parameters across multiple frames With the goal of further improving the computational efficiency.
A final plan involves the investigation of a multi-task training scenario, allowing the model to learn more general transformations.

\bibliographystyle{acm}
\bibliography{biblio}

\end{document}